# A Two-Layer Local Constrained Sparse Coding Method for Fine-Grained Visual Categorization

Guo Lihua*,  Guo Chenggan
South China University of Technology, China, 510641,

Abstract: Fine-grained categories are more difficulty distinguished than generic categories due to the similarity of inter-class and the diversity of intra-class. Therefore, the fine-grained visual categorization (FGVC) is considered as one of challenge problems in computer vision recently. A new feature learning framework, which is based on a two-layer local constrained sparse coding architecture, is proposed in this paper. The two-layer architecture is introduced for learning intermediate-level features, and the local constrained term is applied to guarantee the local smooth of coding coefficients. For extracting more discriminative information, local orientation histograms are the input of sparse coding instead of raw pixels. Moreover, a quick dictionary updating process is derived to further improve the training speed. Two experimental results show that our method achieves 85.29% accuracy on the Oxford 102 flowers dataset and 67.8% accuracy on the CUB-200-2011 bird dataset, and the performance of our framework is highly competitive with existing literatures.

## 1. Introduction

General visual categorization is a highly active research field which has many promising commercial applications. Here, we focus on a relatively smaller but more challenging topic called the fine-grained visual categorization (FGVC) [1]. The FGVC is to classify categories which are both visually and semantically similar, and the goal is to support practical applications such as outdoor plant field guide or animal watching. Now, the FGVC has included many species, i.e. flowers [2], birds [3], dogs [4], trees [5], butterflies [6], and insects [7]. In these species, all categories have the intra-class diversity and inter-class similarity. Therefore, it is very difficult even for humans without professional training to distinguish them.

In the FGVC, local details are more easily distinguished than overall appearances. Intuitively, this character leads us to look into local details. Up to now, lots of successful local descriptors have been successfully used, such as local binary pattern (LBP)[8], part-based one-vs-one features (POOFs)[9], scale invariant feature transform (SIFT)[10] and so on. LBP is powerful when describing the repeated and consistent texture. However it does not suit for the FGVC. POOFs have been proved to be a set of very discriminative intermediate-level features on birds' categorization. However it needs a strong labeled dataset, and requires at least two parts labeled for each image. SIFT quantifies the local gradient orientation, and further forms the bag-of-word (BOW) feature to classify visual images. However, the performance of

BOW feature is less competitive when testing the FGVC [3].

Sparse coding can represent the data efficiently by learning sets of over-complete bases, and a recent work on sparse coding [22], which use Hierarchical Matching Pursuit, outperforms many designed features and algorithms on a variety of recognition datasets. Hierarchical Matching Pursuit looks like an efficient method to recursively extract image features from pixels to patches. However, the discriminative structures may appear at different scales with varying amounts of spatial and appearance invariance. It is necessary for the generic learning model that it can capture the heterogeneity and extract features by recursive sparse coding through many pathways on multiple patches of varying size. Additionally many deep learning approaches [20, 21, 32, 36] have achieved greatly success in the visual classification. Based on these considerations, this paper proposes a two layers local constrained sparse coding method for learning a set of discriminative features, which is shown in Figure 1. The two layers structure can be considered as one kind of deep learning, and final experimental results prove that the two layer structure is most efficient than other deep structures. During sparse coding, an approximate analytical solution is iteratively used to update the dictionary for minimizing the optimization error. Moreover, a local constrained term is introduced into the optimization function for emphasizing the smooth restriction during feature quantification. In sum, our method aims to extract discriminative features as much as possible through the continuous coding process and maximize inter-class differences using the level by level pooling strategy, including the spatial pyramid pooling at final stage.

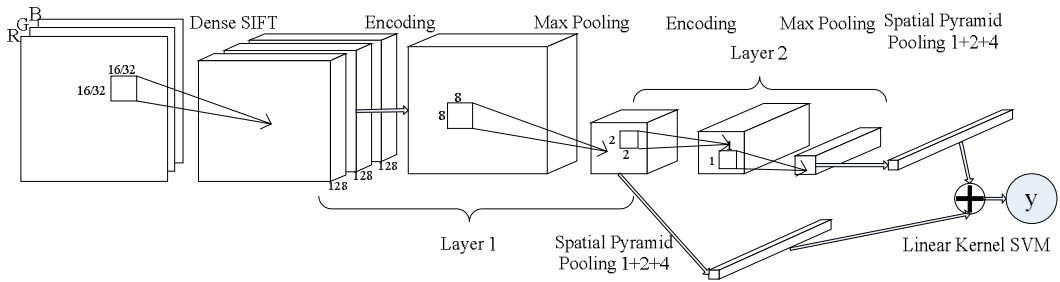

Fig.1 Our two-layer local constrained sparse coding framework. Firstly, Dense SIFT features are computed over color image patches of different sizes (16x16 or 32x32). In layer 1, histogram matrices are encoded into sparse coefficients, followed by a max pooling strategy. Layer 2 does the same procedure. Spatial pyramid pooling is applied at the final stage. Layer 1's features are combined with layer 2's features using a linear kernel, and then train a SVM classifier for classification.

The main process of our method is followed: First, our framework extracts local orientation histograms instead of raw image pixels. Second, two layer architectures are concatenated, and each layer has two components, sparse encoding and max pooling respectively. Third, a spatial pyramid pooling, which aimed to get the discriminative information in spatial distribution, is applied to generate final features. Final, a SVM classifier is used to train and classify the input image.

For evaluating the performance of our method, intensive researches are conducted on two typical and challenging FGVC datasets, i.e. Flowers and Birds. Images of Flowers dataset have symmetrical structures. Comparatively, images of

Birds dataset present high diversity of pose. To cope with the high diversity of bird's poses captured in real world, a pose estimation method for region alignment is necessary to improve the system performance. Our main contributions are summarized as below:

1. A two-layer local constrained sparse coding architecture is applied for the fine-grained visual categorization.
2. A local constrained term is introduced into the optimization goal for guaranteeing the smooth variance when feature quantization and a quick dictionary update.
3. When sparse coding, local orientation histograms take place of raw image pixels for providing more discriminated information.
4. An automatic pose alignment method based on a statistical prior knowledge is proposed.

Next, the related works are reviewed in section 2. More details of our proposed approach are described in section 3 and two experiments are evaluated in section 4. Final conclusions are drawn in section 5.

2. Related Work

Our method on learning a set of discriminative features for FGVC essentially consists of two ideas: aggregating pixels into local spatial information, and coding local orientation information into multi-level local constrained sparse coefficients. Here we first review related work concerning about this feature learning and coding, and then summarize recently fine-grained methods.

In image classification, features are important, and even determine the final performance of classification system [10-18, 38]. In recent years, the most popular approach in visual categorization has been the combining of various descriptors such as SIFT [10], pyramid histogram of visual words (PHOW) [11] or pyramid histogram of oriented gradients (PHOG) [12] and the Bag-of-Visual words [13], which assigning each descriptor into a closest visual codebook entry. Furthermore, there have been several extensions of this popular framework including the use of better designed intermediate-level features based on Fisher kernels such as the vector of locally aggregated descriptors (VLAD) [14] or Fisher vector (FV) [15], and the use of better coding techniques based on soft assignment [16] or sparse coding [17,18]. These features of image will be classified by Support Vector Machine (SVM) [28]. Very recently, Deep networks such as deep belief nets [19], deep auto-encoders [20] and deep convolution neural networks [21] have shown to be powerful in various computer vision tasks. Based on these considerations, varieties of conventional learning and coding frameworks adopt the idea of deep learning while preserving their original rigorous mathematic derivation and have been built into deeper architects to push pixels through multiple layers of feature transform. For example, Bo et al. [22] propose a multipath hierarchical matching pursuit (M-HMP) architecture that combines a collection of hierarchical sparse features for image classification to capture multiple aspects of structures. Following their idea, we find this approach can work better on local orientation histograms. The FGVC is a special case of visual categorization, and has received much progress recently. Berg et al. [9] propose a framework for learning a set of part-based intermediate-level features which require images labeled with part locations. Then in [23] they produce a more practical field guide for a new 500 North American bird species dataset which is publicly available. Chai et al. first propose a bi-level co-segmentation method for image categorization [24] which uses GrabCut algorithm at pixel level and learn an optimal separating hyper-plane on top level. Then Chai et al. propose a tri-level co-segmentation method [25] which minimizes losses at three levels: the category level, the image level and the dataset level. Angelova et al. [26] also propose a detection and segmentation algorithm which first detects low-level regions that could potentially belong to the object and then performs full-object segmentation through propagation. They also collect a larger flower dataset than Oxford 102 flower species dataset [2], which contains 578 different species of flowers and about 250,000 images.

## 3. Proposed Approach

### 3.1. Our Proposed Framework

For an overview, our proposed framework for the fine-grained categorization mainly consists of two stages, fine-grained images alignment and discriminative intermediate-level features extraction, as shown in Figure 2. It is obvious that two kinds of object structure exist in fine-grained images. One kind of structure is a symmetric and static because of limited imaging angle, e.g. flower species, and the other kind of structure is an asymmetric and variable because of moving target, e.g bird species. Therefore pose alignment is very necessary for bird species.

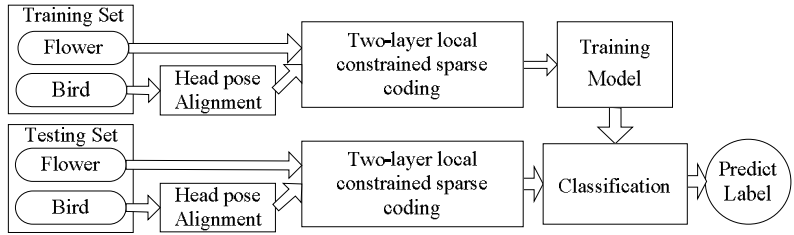

Fig.2 Our proposed framework for fine-grained categorization.

For flower species, we only rescale images into 300 pixels along the shorter side. For bird species, a bird's head part regions are aligned by our automatic pose estimation method. Firstly, a bird's head is detected using the part-based RCNN method. Then the head direction is determined by the voting of several nearest neighbors on training set, and an image is flipped if necessary. Experimental results show that our method only estimates 4 parts with good performance for pose alignment. The detail of experimental result is listed in section 4.2. In two layer stages, dense SIFT features are computed over color image patches of different sizes (16x16 or 32x32). In the layer 1, histogram matrixes are encoded into locally constrained sparse coefficients, followed by a max pooling strategy. The detail of locally constrained sparse coding will be given in section 3.2. The layer 2 does the same procedure. A spatial pyramid pooling is applied to pool valid features. Features from the layer1 and the layer2 are cascaded using a linear kernel, and then a linear one vs. all SVM classifier is trained for classification. Experimental results show that a 3-layer or deeper pipeline is not suitable for our fine-grained categorization method. It is because that it lacks enough samples for dictionary learning process after layer by layer pooling strategy.

3.2. Iterative Learning of codebook using LLC

Sparse coding [18] is to represent signals as a few nonzero entries from a prebuilt codebook. The codebook $D = [d_1, d_2, \cdots, d_M] \in R^{H \times M}$ is designed to be as much redundant as possible in order to sparsely decompose sampled signals $Y = [y_1, y_2, \cdots, y_N] \in R^{H \times N}$ into corresponding sparse codes $X = [x_1, x_2, \cdots, x_N] \in R^{M \times N}$. One standard optimization approach is to minimize the following reconstruction error by forcing codes to be $K$ sparse level

$$\min_{D,X} \|Y - DX\|_F^2, \text{ s.t. } \forall m, \|d_m\|_2 = 1 \ \& \ \forall n, \ \|x_n\|_0 \leq K \tag{1}$$

where $H$, $M$ and $N$ are the dimensionality of the words, the size of the codebook, and the size of training samples, respectively; $\|\cdot\|_F$ denotes the Frobenius norm, $\|\cdot\|_0$ denotes the zero-norm which simply counts the non-zeros entries in the sparse codes $x_n$.

M-HMP [22] considers the possible overfitting problem, and avoids choosing some frequently observed patches with high probability during learning codebooks. Hence they add a regularization term to balance the reconstruction error and the mutual incoherence of the codebook

$$\min_{D,X} \|Y - DX\|_F^2 + \lambda \sum_{i=1}^{M} \sum_{j=1, j \neq i}^{M} |d_i^T d_j|, \text{ s.t. } \forall m, \|d_m\|_2 = 1 \ \& \ \forall n, \ \|x_n\|_0 \leq K \tag{2}$$

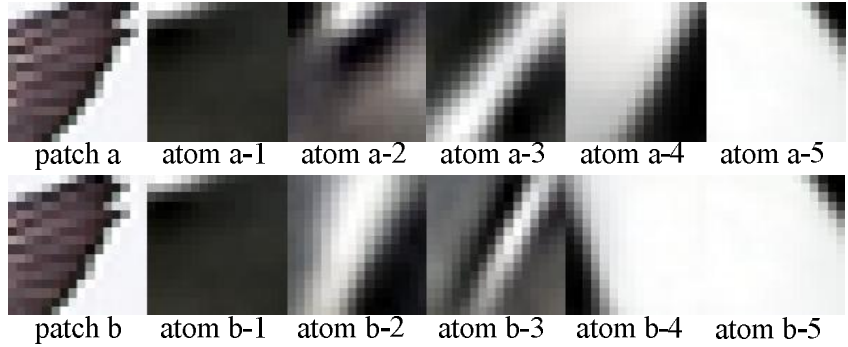

patch a   atom a-1   atom a-2   atom a-3   atom a-4   atom a-5

patch b   atom b-1   atom b-2   atom b-3   atom b-4   atom b-5

Fig.3 Image patches and their corresponding selected atoms.

However, in our experiments, image patches with similar visual pattern are found to select quite different visual words when using M-HMP method. As is shown in Figure 3, the patch A is similar to the patch B, but they only share 2 same atoms while rest 3 atoms are very visually different. The noise of image patches will cause the dictionary learning of M-HMP unstable. An ideally updating process is that a smooth constraint is emphasized during the learning process. It means that those closely distributed dictionary atoms are necessarily selected by visually similar patches. To incorporate a local smooth property and fast encoding process, a third regularization term is added on the M-HMP object function that can be iteratively optimized. The main merit is that it takes advantage of both the convenience of approximated analytical solution derived in LLC [18] and K-SVD [29] dictionary learning algorithm. The learning process is described in details as a kind of pseudo-code in Table 1. Our optimization function is following:

$$\min_{D,X} \|Y - DX\|_F^2 + \lambda \sum_{i=1}^{M} \sum_{j=1, j \neq i}^{M} |d_i^T d_j| + \beta \sum_{i=1}^{N} \|e_i \square x_i\|^2 \quad (3)$$

$$s.t. \ \forall m, \|d_m\|_2 = 1 \ \& \ \forall n, \ 1^T x_n = 1$$

where $e_i = \exp([dist(y_i, d_1), \cdots, dist(y_i, d_M)]^T / \sigma)$, $dist(y_i, d_j)$ is the Euclidean distance between $y_i$ and $d_j$, $\square$ denotes the element-wise multiplication, and $\sigma$ is a weight adaptor. The iterative optimization is used to solve the equation 3, and the details are below:

**First, encoding:** Given a codebook $D$ to calculate the sparse codes $X$, equation (3) is formed into the classical LLC object function (4) which has an analytical solution (5)

$$\min_{D,X} \|Y - DX\|_F^2 + \beta \sum_{i=1}^{N} \|e_i \square x_i\|^2 \quad (4)$$

$$s.t. \ \forall n, \ 1^T x_n = 1$$

$$\begin{cases} C_i = (D^T - 1y_i^T)(D^T - 1y_i^T)^T \\ \tilde{x}_i = (C_i + \beta diag^2(e)) \backslash 1 \\ x_i = \tilde{x}_i / 1^T \tilde{x}_i \end{cases} \quad (5)$$

Practically we simply use the approximated LLC of original paper [18] to speed up the encoding process. Since we iteratively minimize our object function, the loss in the approximated encoding can be worthy comparing to computation complexity.

**Second, codebook update:** Given the sparse codes $X$, the codeword $d_m$ can be optimized sequentially using K-SVD algorithm. In the m-th codebook update step, we remove the constant terms in (3), thus we have

$$\min_{d_m} \left\{ \bar{x}_m^T \bar{x}_m d_m^T d_m - 2R_m \bar{x}_m + \lambda \sum_{j=1, j \neq m}^{M} |d_j^T d_m| + \beta \sum_{i=1}^{N} \exp(2dist(y_i, d_m) / \sigma) * \bar{x}_{mi}^2 * \delta_{mi} \right\} \quad (6)$$

$$s.t. \|d_m\|_2 = 1$$

where $\bar{x}_m^T$ is the row of $X$, and $R_m = Y - \sum_{i \neq m} d_i \bar{x}_i^T$ is the residual matrix for the m-th codeword, $\bar{x}_{mi}$ is $(m,i)$ element in $X$, $\delta_{mi} = 1$ if $\bar{x}_{mi} \neq 0$, otherwise $\delta_{mi} = 0$. To solve (6), we find both $0 < \exp(2dist(y_i, d_m) / \sigma) < 1$ and $0 \leq |\bar{x}_{mi}| < 1$ making third term almost zero, so we may leave out the third term in practice. From another point of view, the third term is introduced to enforce local constrains during encoding process, and keeping the third term in codebook update process just makes $d_m$ closer to training samples.

In sum, sparse coding coefficient can be solved by an analytical solution, and iterative optimization is used to solve our optimization equation 3. As we all known, these two models can guarantee the convergence of system. It is shown in Figure 5 that the learning process quickly comes to the convergence point after a few iterations. In all our experiments, the iteration number is only 10.

Tabel.1 The pseudo-code of proposed iteratively codebook learning method

**Initial input**: L2-normalized overcomplete DCT codebook $D^1 = [d_1, d_2, \cdots, d_M] \in R^{H \times M}$; Sampled descriptors $Y = [y_1, y_2, \cdots, y_N] \in R^{H \times N}$; Iterations number $I$; Sparse level $K$.

**Loop $i$ from 1 to $I$:**

- Encoding, parallel loop $n$ from 1 to $N$:

$$dis_m = \|y_n\|_2^2 - 2y_n^T d_m^i, \forall m \in M,$$

$$idx = sort(dis_1, \ldots, dis_M, 'ascend'),$$

$$z_k = d_{idx_k}^i - y_n, \forall k \in K,$$

$$c = z^T z + eye(K, K) * \beta * trace(z^T z),$$

$$w = c \backslash 1, 1^T w = 1, \|w\|_2 = 1,$$

$$\begin{cases} X_{idx_k,n}^i = w_k, \forall k \in K, idx_k \in M \\ X_{m,n}^i = 0, \quad \forall m \in M, m \neq idx_k \end{cases},$$

- Codebook Update, loop $m$ from 1 to $M$:

$$idx = \{idx_1, \ldots, idx_J\}, \forall j \in J, X_{m,idx_j}^i \neq 0,$$

$$X_{m,}^i = X_{m,idx}^i, X' = X_{,idx}^i,$$

$$[d_m^{i+1}, s, X_{m,}^i] = svds(Y_{,idx} - D^i X' + d_m^i, 1),$$

$$\begin{cases} X_{m,idx}^{i+1} = s X_{m,}^i \\ d_m^{i+1} = \dfrac{d_m^{i+1}}{\|d_m^{i+1}\|_2} \end{cases},$$

For every $m \in M$, replaced $d_m^{i+1}$ with a normalized descriptor $y_n = \dfrac{y_n}{\|y_n\|_2}$ if $\forall j \neq m, \max(|d_j^T d_m^{i+1}|) > muthresh$, and *muthresh* was a mutual incoherence limit.

**Output**: Codebook $D^I$ and corresponding sparse codes $X^I$

## 3.3. Local Orientation Histogram Rather Than Pixels

We believe that all information needed to classify fine-grained categories lies in pixels. However, pixels' values are easily affected by various environmental factors. In our methods, the local orientation histogram takes place of the raw pixels. We transform pixels of local regions into orientation histograms to make the description robust to intra-class and inter-class variations. Specifically, most common examples are shifted, rotated, and scaled images. However, from local view, object edges in local regions mainly remain unchanged and show repeated patterns. Hence, we find it necessary to adopt a histogram quantization procedure like SIFT before learning sparse dictionaries. We calculate gradient maps on R, G, and B channels of a 16x16 image patch respectively, and then quantify them into orientation histograms, as is shown in Figure 4. Practically the size of image patches may be different. To handle different scales of images, we simply use linear kernel to combine several image patch sizes as a multi-scale approach. For all the following experiments, we set the stride size as one pixel to maximize the possible pattern combinations for coding. From another perspective, our processing looks like the convolution kernel operation, and transform the pixels of local regions into a representative feature space.

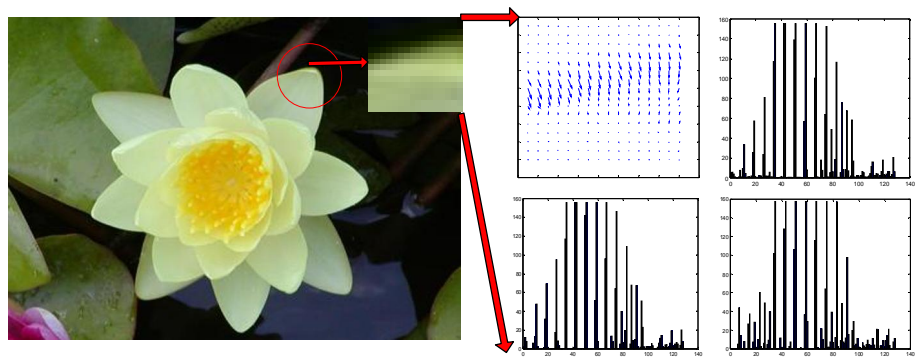

Fig.4 Workflow of calculating local orientation histograms on R, G, B channel, respectively.

The main merit is that this description is more robust, and makes recognition algorithms converge quickly. As is shown in Figure 5, the RMSE of using local orientation histogram is much lower than that of directly using pixels in the iteration of learning codebook.

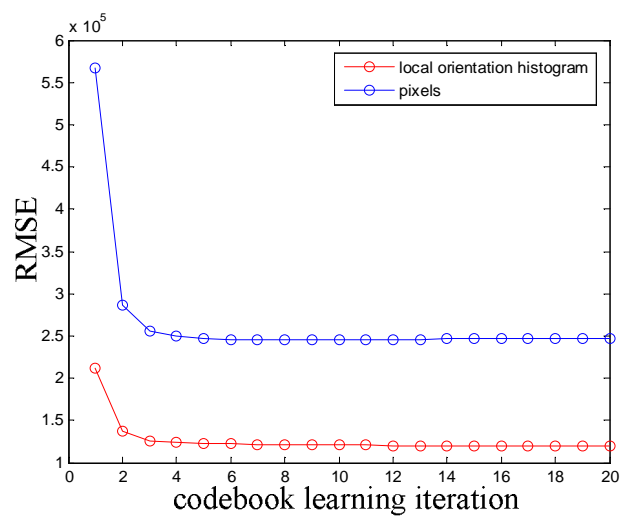

Fig.5 The RMSE between the reconstructed descriptors and the original descriptors after every updating iteration. The red line means using local orientation histograms and the blue line means directly using pixels.

3.4. Part Estimation for Bird Species based on Statistical Method

Recently, many part detection methods for bird species have been developed, such as part-based R-CNNs [32], Nonparametric Part Transfer [33], Pose Pooling

Kernels [34] and so on. Here, we propose an automatic part estimation method based on the statistical information of training images, and the framework is shown in Figure 6.

Firstly the part-based R-CNNs method is used to roughly detect a bird's head in a given image. From the view of bird, a bird's head direction can be decided based on the combination of the bird's head and beak. In our method, a bird's head direction is modeled as three directions, i.e. heading left, middle or right. In the training dataset, the location of a bird's head and beak is given. Therefore, the head direction can be gotten by the relative relation between the bird's head and beak. As for the testing dataset, the head direction of a testing image can be calculated as following:

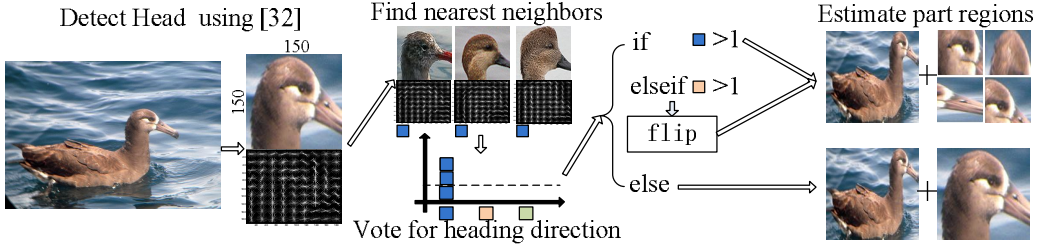

Fig.6 Our part estimation flow work for bird species. ■ means heading right, ■ means heading left and ■ means heading middle.

First, we calculate HOG descriptors for the head region in a test image, and then choose three candidate training images which are most visually similar in HOG descriptors. In the training dataset, the head and beak location are given, and we classify the training image into three directions, i.e. heading left, middle or right. Therefore, the test image can be determined to heading left, middle or right by pulling votes of candidate directions of three training images. After deciding the head direction of training and testing image, the training and testing image must be a unified framework. If the image is heading left, it will be flipped to be heading right, and then we generate features from four parts of head, i.e. eye, beak, forehead and crown. If the image is heading right, we only generate features from four parts of head. If the train or test image is heading middle, it only needs to be recorded, and generate features directly on head regions instead of four parts of head.

Second, in the training dataset, the four head part coordinates must be normalized excluding those images with heading middle. We calculate the ratio between the coordinates of a bird's head part and its corresponding head bounding box as following:

$$(ratio^i_{part\_x}, ratio^i_{part\_y}) = (\frac{x^i_{part} - x^i_{bbox}}{width^i_{bbox}}, \frac{y^i_{part} - y^i_{bbox}}{height^i_{bbox}}), \quad (7)$$

$$\forall i \in training\ set, part \in \{eye, beak, forehead, crown\}$$

where $(x^i_{part}, y^i_{part})$ is the labeled part location in the $i$-th image, and $(x^i_{bbox}, y^i_{bbox}, width^i_{bbox}, height^i_{bbox})$ is the location and size of head bounding box in the *i*-th image. The statistical distribution of four parts coordinates on training set is plotted in Figure 7. The top left image in Figure 7 shows that the distribution of four parts coordinates looks like the Gaussian distribution, and a clear boundary can be draw between different parts. For example, the eye of bird mainly stays in the middle of the head bounding box if the bird is heading right. This kind of uniform distribution also matches our intuitive knowledge in daily life. The other images in Figure 7 show that the histogram of each part coordinate also looks like the Gaussian distribution. During calculate the histogram of each part coordinates, our method uses 150 bins because all image are rescaled into 150*150 sizes. Therefore, we fit Gaussian probability density to match the statistical data of each part. For x and y coordinate respectively, we have

$$f(x) = \frac{1}{\sqrt{2\pi}\sigma} e^{-(\frac{x-\mu}{\sqrt{2}\sigma})^2}, f(y) = \frac{1}{\sqrt{2\pi}\sigma} e^{-(\frac{y-\mu}{\sqrt{2}\sigma})^2} \qquad (8)$$

where $\mu$ means the mean and $\sigma$ means the standard deviation in observed samples.

We adopt the $3\sigma$ principle to cover 99% of a possible part's region. For example, in our experiments the proposed right eye's region for right pose is following:

$$x: \mu \pm 3\sigma = 78 \pm 44 \, pixels$$
$$y: \mu \pm 3\sigma = 63 \pm 34 \, pixels \qquad (9)$$

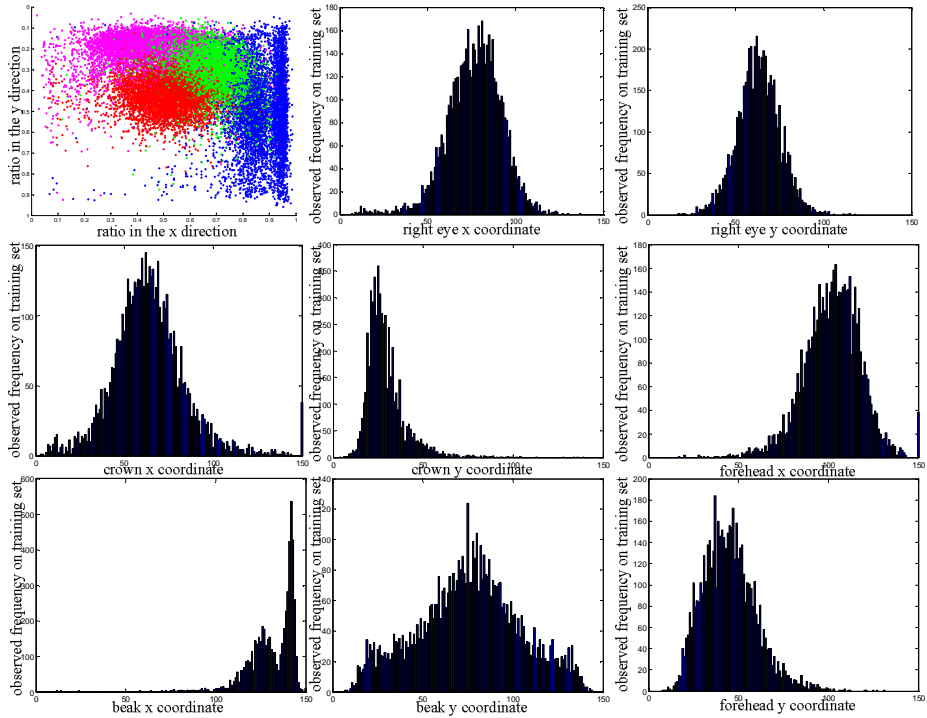

Fig.7 Top Left: Statistical distribution of four parts coordinates on training set, eye's points are given in red color, beak's points are in blue color, forehead's points are in green color and crown's points are in magenta color. From Top Middle to Bottom Right: Four parts corresponding coordinate's histograms with 150 bins, all these histograms can be fitted by Gaussian curves.

The core idea behind our method is that we can always detect a bird's head in an image captured from an alive bird, and the functional parts' arrangement remains unchanged for all those categories. It is relatively predictable and stable that different bird categories share the same biological parts. Figure 8 shows five examples on proposed potential part regions for testing set.

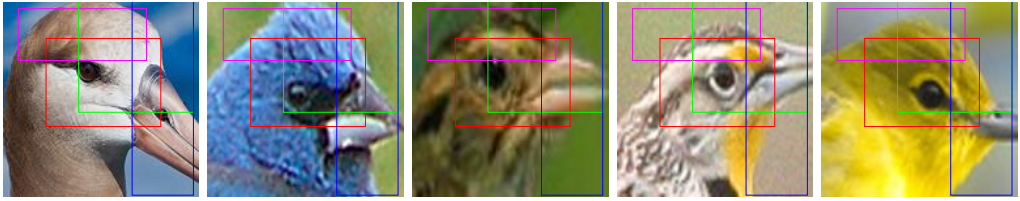

Fig.8 Proposed part regions on testing set. Eye's region is given in red color, beak's region is in blue color, forehead's region is in green color and crown's region is in magenta color.

## 4. Experiments

Two public fine-grained datasets, i.e. the Oxford 102 Flowers dataset [2] and the Caltech 200 Birds (2011) dataset [3], are evaluated our method with some existing works.

The flower dataset contains 8189 images of flowers belonging to 102 different categories. For each category, 10 images are used for training, 10 for validation, and the rest for testing, as the same as in [2]. We roughly crop all the images with a bounding box and rescale them into 300 pixels along the shorter edge. The bird dataset contains 200 bird species. For this dataset, we follow the suggested train/test split setting, as the same as in [3]. Image numbers (#) of two dataset in the training, validation and testing set are listed in Table 2. We also crop images using the provided bounding box, and extract the detected part regions by our pose estimation method, which are rescaled into 150x150 pixels. Samples from these two datasets are shown in Figure 9.

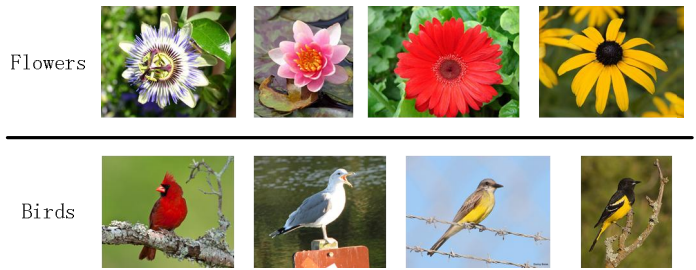

Fig.9 Examples from FGVC datasets, namely Oxford 102 flowers and Caltech 200 birds, respectively.

Table 2. Image numbers (#) of two dataset in the training, validation and testing set

| Dataset | training | validation | testing |
| --- | --- | --- | --- |
| Flowers | 1020 | 1020 | 6194 |
| Birds | 5994 | no | 5794 |

### 4.1. Experiment of Oxford 102 Flowers

Before the comparative experiment, we must previously fix important parameters. As a set of default parameters, the codebook size is 2000, the image patch size of extracting local orientation histograms is 16x16, the patch size of max pooling is 4x4, and the sparse level is 4. Then we only change one of these parameters to evaluate the performance on validation set while simultaneously fixing the others. A linear SVM is our classifier. In our experiments, we random select 200 feature samples on each training image, and the dimension of each feature sample is 384 because of the concatenation of dense SIFT descriptors on R, G and B channels. In the validation set, this leads to a set of 20400 feature samples for codebook training.

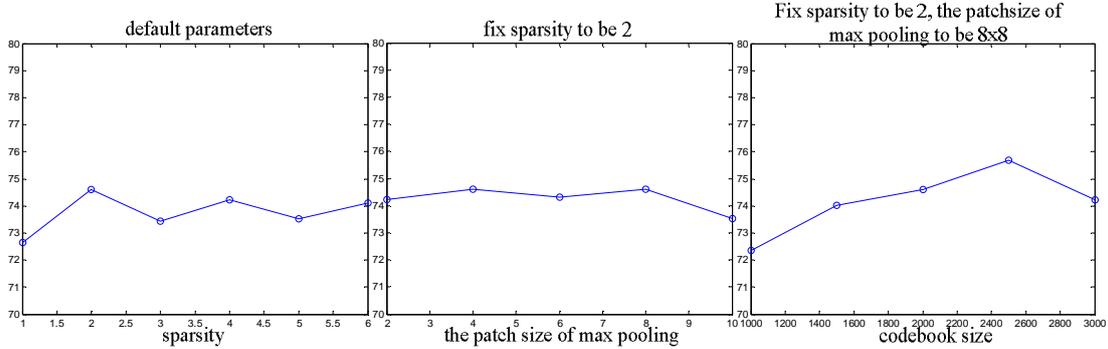

Fig.10 Left: the average accuracy of different sparse level, other parameters remains default. Middle: the average accuracy of different max pooling patch size, while fixing sparse level to be 2 and others to be default. Right: the average accuracy of different codebook size, while fixing max pooling patch size to be 8, sparse level to be 2, and others to be default.

The left image in Figure 10 shows that the performance is best when sparse level is 2. Sparse level 1 means that the sparse coding degenerates into a bag of visual word approach with a hard encode technique. Sparse level 2-6 means that the sparse coding increases local entries for encoding features. With the sparse level rising, the accuracy first increases for about 2%, and then fluctuates. By considering the encoding efficiency and classification performance, we fix the sparse level to be 2 in all our experiments. Next we vary the patch size of max pooling to check how this parameter affects final performance. The middle image in Figure 10 shows that the accuracy is relatively flat until the patch size becomes large than 8. So we fix the max pooling patch size to be 8x8. To a codebook based method, the size of codebook plays an important role in classification. The codebook size is set to be 2500 after cross-validation, as shown in the right image of Figure 10.

After parameter evaluation, we summarize our parameters in next experiments, which show in Table 3. There are two pipelines existing in our method, i.e. shallow single layer architecture and deep two-layer architecture. For both pipelines, a set of denseSIFT features with a stride of 1 pixel are extracted on 16x16 or 32x32 RGB image patches to preserve the color information. This makes our input feature samples to be 384 dimensions. In single-layer architecture setting, our dictionary size is 2500, max pooling patch size is 8 and sparsity level is 2. In two-layer architecture setting, the first layer dictionary size is 800, pooling size is 8 and sparsity level is 2; the second layer dictionary size is 4000, pooling size is 1 and sparsity level is 8. Both pipelines are followed by a spatial pyramid pooling strategy with a level of 1, 2, and 4. It's worth noting that dictionaries in both pipelines and scales are trained separately to best fit for features. Compared with other nonlinear kernel type, linear kernel is the best performance, which is shown in Table 4. Therefore, at the final stage, a linear kernel is adopted to train a one-vs-all SVM classifier.

Table.3 the summarization of our parameters, A and B refer to different image patch sizes.

|  |  | A1 | B1 | A2 | B2 |
|---|---|---|---|---|---|
| Patch size | | 16x16 | 32x32 | 16x16 | 32x32 |
| spatial pyramid pooling | | Yes | Yes | Yes | Yes |
| Layer | | 1 | 1 | 2 | 2 |
| Layer 1 | dictionary size | 2500 | 2500 | 800 | 800 |
| | sparsity level | 2 | 2 | 2 | 2 |
| | max pooling patch size | 8 | 8 | 8 | 8 |
| Layer 2 | dictionary size | - | - | 4000 | 4000 |
| | sparsity level | - | - | 8 | 8 |
| | max pooling patch size | - | - | 1 | 1 |

To compare with the M-HMP method, we directly use the parameters given from the paper [22] to test on the Oxford 102 flowers dataset. Table 5 shows our recognition accuracies. The experimental results show that the recognition accuracy of a single 2-layer pipeline doesn't outperform a single 1-layer pipeline. However, by combining these two pipelines, we really can improve accuracy. The M-HMP method includes a 3-layer pipeline, however, the dataset lacks enough samples for dictionary learning process after layer by layer pooling strategy, our method only includes 2-layer pipeline.

Table.4 Recognition accuracies on choosing different SVM kernel types

| Kernel Type | Linear | Polynomial | RBF | Intersection |
|---|---|---|---|---|
| Accuracy | 85.29 | 83.43 | 84.12 | 84.61 |

Table.5 Average recognition accuracies(%) on Oxford 102 flowers.

|  | Ours | M-HMP |
|---|---|---|
| A1 | 84.51 | 67.25 |
| B1 | 82.94 | 64.8 |
| A2 | 80.78 | 59.41 |
| B2 | 80 | 62.16 |
| B3 | - | 59.22 |
| all | 85.29 | 78.43 |

In Table 6, we compare our method with six recently published algorithms: PCANet [27] (simple version of deep neural network), Nilsback's method [2], Bicos[24], BicosMT[24], Angelova's method [26] and saliency method [31]. PCANet is a baseline of deep neural network, and is validated that it is efficient in the face recognition. The parameters of PCANet is followed: the stages is 2, the patch size is 3x3, the number of filters for each stage are 8 and 8, the block size of histogram for each stage are 7 and 7, the ratio of overlap is 0.7, and the images are 50x50 pixel on R, G, B channel. Table 6 shows that the performance of our method is better than that of deep neural network (PCANet). Generally, methods with segmentation will perform better than those methods without segmentation. However, our method does not adopt those segmentation methods here because our primary purpose is to propose an efficient feature extracting method. Experimental results in Table 6 show that our method has a higher accuracy than those segmentation methods [24, 26]. The average accuracy of our method is 85.29% accuracy, which is the best.

Table.6 Recognition performance on Oxford 102 flowers.

Methods including a segmentation procedure are noted as "seg", otherwise noted as "no-seg".

| Methods | Average Accuracy(%) | Training set # | segmentation |
|---|---|---|---|
| Chan et al. PCANet ([27]) | 68.28 | 2040 | no-seg |
| Nilsback and Zisserman ([2]) | 72.8 | 2040 | no-seg |
| Chai et al. Bicos ([24]) | 79.4 | 2040 | seg |
| Chai et al. BicosMT ([24]) | 80 | 2040 | seg |
| Angelova and Zhu( [26]) | 80.66 | 2040 | seg |
| Hu et al. ( [31]) | 81.51 | 2040 | no-seg |
| **Ours** | **85.29** | 2040 | no-seg |

4.2. Experiment of Caltech 200 Birds dataset

After the comparative experiment, best parameters are the same with those in Flowers datasets, which is listed in Table 3.

First, we evaluate the performance of our proposed head direction voting method for part estimation. To set up a groundtruth set, images with right eye visible are considered to be heading right direction, with left eye visible considered to be heading left direction and both eye visible considered to be heading middle direction. By choosing 1, 3, 5, 7 neighbors, accuracies are plotted in Figure 11. Left image in Figure 11 shows that the accuracy is about 50% when using 1 candidate training image; however the accuracy is about 90% when using 3 or more candidate training images. Based on this, the number of candidates is chosen as 3 in our experiments.

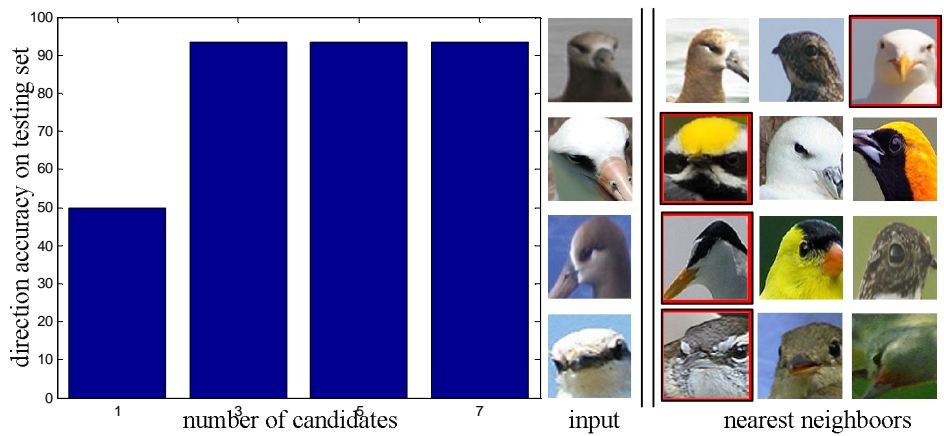

Fig.11 Left: the heading direction accuracy of different number of candidates. Right: some failed voting cases with 3 candidates.

Even though this voting method has a high accuracy, the right column of Figure 11 shows some failed voting cases to illustrate the possible mismatching situation. A simple nearest search strategy adopted here may fail to match some complicate appearances. However, our method can extract discriminative features through the continuous coding process and using the level by level pooling strategy. Therefore, precisely location of the unknown part in a given test image is not necessary to our method.

Second，we evaluate the performance of different part in head. The experiment is conducted by only using the single-layer pipeline with a patchsize of 32x32. All part regions are centered at the given locations with a region size of 1/16 proportion image size, then the extracted part regions are rescaled into 150x150 pixels. Figure 12 shows the appearance frequencies and classification accuracies of each part. The red bars are the normalized classification accuracies of each part, and the blue bars are the normalized appearance frequencies of each part. Appearance frequencies represent the most common parts that can be seen in an image. From the stacked bars in Figure 12, the stacked bars of eye, beak, forehead and crown are higher than those of others. These four parts are around a bird's head, and contribute a lot for bird species classification. It can also explain why we chose four parts as our head part detection method.

Third, to be consistent with the experiments on flower dataset, Table 7 also gives the comparison between our method and the M-HMP method. It should be noted that the original M-HMP method hasn't given experimental results on Caltech 200 birds, so we only use the default experiment settings that only test classification accuracy on provided object bounding box. But our method includes a head part region alignment process designed according to the characteristics of bird images. From the classification accuracy in Table 7 it can be draw a similar conclusion as the experiments on flower dataset. In the meantime it also explains that for bird dataset an image alignment process can largely improve the recognition performance.

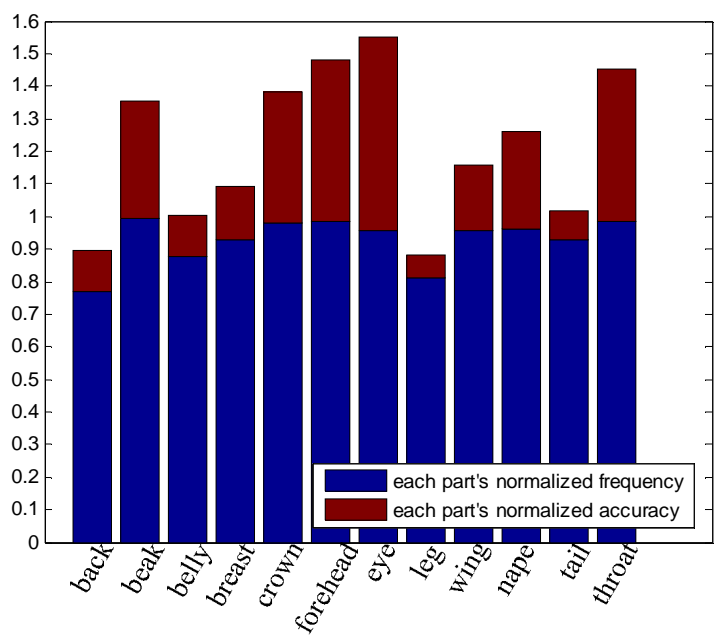

Fig.12 Red bar: the normalized classification accuracy of proposed method on each part region. Blue bar: the normalized appearance frequency of each part on the bird species dataset.

Table.7 Average recognition accuracies (%) on Caltech 200 birds.

|     | Ours(without alignment) | Ours  | M-HMP |
|-----|------------------------|-------|-------|
| A1  | 39.87                  | 64.24 | 46.39 |
| B1  | 35.73                  | 65.95 | 37.59 |
| A2  | 30.12                  | 60.62 | 29.76 |
| B2  | 27.25                  | 61.02 | 31.44 |
| B3  | -                      | -     | 26.65 |
| all | 40.35                  | **67.80** | 48.41 |

Table 8 shows the comparative performance between our method with existing literatures, i.e. POOF[9], NPT [33], Alignments [35], deep learning methods like Part-based RCNN [32] and DPD+DeCAF [36], and Symbiotic Segmentation [37]. Experimental results show that the performance of our method without pose alignment is the worst, which means the pose alignment is necessary to Birds dataset. After our automatic pose alignment, the performance of our method is better than many reported methods except for the latest part-based RCNN approach [32]. Moreover, our part region estimation process doesn't need to precisely locate parts. And nearest neighbors are only searched for classifying heading direction. Therefore, our method doesn't require any part locations on testing set while POOF [9] needs. In NPT method [33], its performance highly depends on the precisely location transform from training set, however the location transform is not robust. Preprocessing of object segmentation is helpful in the fine-grained categorization. However, the final classification accuracy of our method is about 8.4% higher than Symbiotic Segmentation method [37]. Though the part-based RCNN method [32] seems to be much better than our method, it needs to be pre-trained on the large ImageNet dataset, whose image numbers is about 14 millions, before fine-tuning for bird species. Therefore the part-based RCNN method [32] needs the extra dataset information.

Table 8. Recognition performance on Caltech 200 birds

| Methods | Average Accuracy(%) | Total # images of Training set | Pose alignment |
|---------|---------------------|-------------------------------|----------------|
| POOF [9] | 56.78 | Birds(5994) | need |
| NPT [33] | 57.84 | Birds(5994) | need |
| Symbiotic Segmentation [37] | 59.4 | Birds(5994) | need |
| Alignments [35] | 62.7 | Birds(5994) | need |
| DPD+DeCAF [36] | 64.96 | Birds(5994) | need |
| Part-based RCNN [32] | 76.37 | ImageNet(14Millions)+Birds(5994) | need |
| Ours(without alignment) | 40.35 | Birds(5994) | No |
| **Ours (alignment)** | **67.8** | Birds(5994) | need |

## 5. Conclusions

A two-layer local constrained sparse coding architecture is proposed to solve the fine-grained visual categorization. The two-layer architecture is to learn intermediate-level features, and the local constrained term is to guarantee the local smooth of coding coefficient. During sparse coding, local orientation histograms take place of raw pixels for extracting more discriminative information. To further improve the training speed, a quick dictionary updating process is derived. Moreover, when evaluating the bird dataset, we propose a pose estimation method for region alignment to cope with the high variety of bird's poses. Classification performance is verified on two public fine-grained datasets named Oxford 102 flowers and Caltech 200 birds. Our method achieves 85.29% accuracy on the Oxford 102 flowers dataset and 67.8% accuracy on the CUB-200-2011 bird dataset, which is highly competitive with existing literatures. Due to the limitation of dataset's sample for dictionary learning process after the layer by layer pooling strategy, the two layer structure is most efficient than the other deep structures, which has been proved by final experimental results. In the future, the large scale fine-grained category dataset may be collected, and the deeper structure may be intensively evaluated in the large scale dataset. Moreover, a three level spatial pyramid pooling is applied, and the dimension of output features is high, which need to occupy a large memory space. In our future work, feature dimension reduction methods or dictionary compression methods will be incorporated in our learning framework to further reduce memory usage and improve computing efficiency.